\newcommand{\CLASSINPUTbottomtextmargin}{2.5cm}
\documentclass[10pt,conference,a4paper]{IEEEtran}
\IEEEoverridecommandlockouts
\usepackage{cite}
\usepackage{amsmath,amssymb,amsfonts}
\usepackage{algorithmic}
\usepackage{graphicx}
\usepackage{textcomp}
\usepackage{xcolor}
\def\BibTeX{{\rm B\kern-.05em{\sc i\kern-.025em b}\kern-.08em
    T\kern-.1667em\lower.7ex\hbox{E}\kern-.125emX}}

\begin{document}

\title{Deep Learning for Assessment of Oral Reading Fluency}

\author{\IEEEauthorblockN{Mithilesh Vaidya, Binaya Kumar Sahoo, Preeti Rao\thanks{This paper reports work undertaken at IIT Bombay in 2022.}}
\IEEEauthorblockA{\textit{Electrical Engineering} \\
\textit{Indian Institute of Technology Bombay}\\
Mumbai, India}

}

\maketitle

\begin{abstract}
Reading fluency assessment is a critical component of literacy programmes, serving to guide and monitor early education interventions. Given the resource intensive nature of the exercise when conducted by teachers, the development of automatic tools that can operate on audio recordings of oral reading is attractive as an objective and highly scalable solution. Multiple complex aspects such as accuracy, rate and expressiveness underlie human judgements of reading fluency.  In this work, we investigate end-to-end modeling on a training dataset of children's audio recordings of story texts labeled by human experts. The pre-trained wav2vec2.0 model is adopted due its potential to alleviate the challenges from the limited amount of labeled data. We report the performance of a number of system variations on the relevant measures, and also probe the learned embeddings for lexical and acoustic-prosodic features known to be important to the perception of reading fluency.   

\end{abstract}


\section{INTRODUCTION}\label{sec:intro}

Reading is a foundational skill in today's world. Children who are poor readers suffer in school and in life at large~\cite{aser}, ~\cite{torgesen1998catch}.  The identification of poor readers can help draw attention in terms of pedagogic interventions. A framework which can reliably assess the reading skills of children at scale therefore has the potential for massive social impact. With comprehension being the ultimate goal of reading, reading fluency has been typically assessed by question answering tasks. However, a more direct and cognitively less demanding approach is to evaluate fluency via the reading aloud of grade-level texts. Reading research indicates that comprehension is demonstrated through the reader's ease of word recognition, appropriate pacing and expression~\cite{kuhn2010aligning}.  
Thus, fluency can be measured by the accuracy, automaticity, and prosody demonstrated in oral reading, all of which facilitate the reader’s construction of meaning.

Word decoding ability is judged by measuring accuracy (number of words read correctly) and speed, combining these to obtain a metric called WCPM (words correct per minute)~\cite{hasbrouck1992curriculum}.  Since word decoding ability is the first stage in comprehending text, readers with low WCPM scores can be immediately flagged for poor reading comprehension.  Once a child attains a certain level of word decoding proficiency, prosodic aspects start gaining more importance~\cite{valencia2010oral}. In order to comprehend the text and make it comprehensible to a listener, the reader needs to identify the words, relate them to words in the neighbourhood by taking the sentence structure and syntax into account and update an internal mental model. When reading text aloud, we modulate our voice to reflect this internal process. It is acoustically correlated with prosodic aspects of speech such as rhythm, pausing, intonation and stress. Proficient readers tend to convey the linguistic focus and the author's sentiments using prosodic variations. As a result, a simple ASR-based framework which compares the ASR-decoded text with the intended text will be an unreliable indicator of oral reading ability since it fails to capture the all important prosodic aspect. Thus, high WCPM is a necessary but not a sufficient criterion for reading proficiency.


With relatively limited research in automated prosody modeling for reading, recent work proposed a large set of hand-crafted features was derived from the waveform, ASR decoded output and the reference or canonical text for the comprehensibility prediction of short passages of read aloud text ~\cite{kamini_thesis, sabu2024predicting}. Acoustic features include functionals of pitch, intensity and energy at various hierarchies (word-level and recording-level) while lexical features include word-level POS tags, miscue features, etc. and recording-level features such as WCPM and prosodic miscue aggregates. Recursive Feature Elimination using Cross Validation (RFECV) is used in tandem with a Random Forest Classifier (RFC) to derive a compact set of features of highly relevant features. The best system, operating on features derived from the waveform and manually transcribed text, reported a Pearson corr. of 0.794 with manual scores, while also demonstrating the significant contribution of prosody features on system performance. However, such a system has a few drawbacks. Firstly, the feature extraction stage has multiple independent components which cannot be jointly optimised for the task of fluency prediction. Secondly, extracting such features requires significant domain knowledge. Thirdly, since the feature space is very large, we might miss out on important features or patterns in the data when using hand-engineering.

With the advent of deep learning, end-to-end models trained on vast datasets have outperformed traditional methods in almost every domain. Their key promise is the ability to learn powerful representations directly from data. Zhou et al.~\cite{yu2015using} proposed a deep learning model for fluency assessment of non-native English speakers. The last time step output of a bidirectional LSTM, operating on frame-level contours (such as F0, loudness, MFCCs, etc.) is extracted. It is concatenated with utterance-level hand-crafted features (such as fluency, grammar and vocabulary use) and passed through a simple linear layer to predict a 4-point score for 45 seconds of spontaneous speech. They observed an improvement of around 1\% improvement over the use of hand-crafted features. An improved architecture was subsequently proposed by the same group in~\cite{chen2018end}. Word-level acoustic features (such as mean pitch) and lexical features (GloVe embeddings of ASR hypothesis) are input to feature extractors such as CNNs and attention-augmented RNNs. The output of each modality is concatenated and passed through a linear regression layer to predict the score. They observed that BLSTM, coupled with an attention layer, significantly improved performance.  In another work, outputs of the acoustic branch, consisting of a CRNN operating on spectrograms, and a BLSTM, operating on GLoVe embeddings, are fused with a novel attention fusion network~\cite{grover}. The attention weights can be interpreted as the importance of each modality in predicting the final score. Lastly, separate BLSTMs are proposed in~\cite{qian_scoring} for modelling Delivery, Language use, and Content in an automated speech scoring system. Each response in the dialogue is rated separately by the three sequential models. The final speech proficiency score is obtained by fusing the three subscores. Their proposed system attained a Pearson corr. of 0.747, an improvement of 0.063 over the RFC baseline operating on hand-crafted speech features~\cite{zechner2009automatic}. Note that all the above systems operate on some set of hand-crafted features, be it at the frame, word or recording level.

In this work, we investigate an end-to-end deep learning framework for automatic reading fluency assessment. Our model is based on Wav2vec2.0~\cite{wav2vec} (referred to as wav2vec hereafter). Pre-trained on a large unlabelled speech corpus in a self-supervised fashion, wav2vec generates a robust frame-level representation of speech. In this work, we use a pre-trained model to extract frame-level features and experiment with various deep learning modules on top to predict the comprehensibility rating. We are also interested in how the different pre-trained models that are publicly available affect system performance on our task.  

A drawback of deep learning systems is their black-box nature. Despite superior performance over traditional techniques, their internal representations are opaque. This presents obstacles in certain applications and also hinders further improvement possible through identifiable complementary information. 
Some recent works~\cite{pasad2021layer, shah2021all} have probed transformer representations for features such as phone identity and acoustic contours.  We try to emulate this for our task via correlations of the embeddings with interpretable features known to matter in the perception of fluency or comprehensibility.


\section{DATASET AND TASK} \label{sec:dataset}

We use an available dataset of audio recordings by children belonging to the age group of 10-14 years with English as the second language studied in school~\cite{kamini_thesis,sabu2024predicting}. 
The 1447 recordings in the dataset are manually transcribed and pre-screened for WCPM$>$70. We have 165 speakers reading from a pool of 148 unique paragraph texts. Each paragraph has 50-70 words. The duration of the recordings varies from 12 seconds to 56 seconds with a mean recording duration of 25 seconds and standard deviation of 8 seconds. The total duration of the dataset is approximately 10 hours. 

Each recording is independently rated for 'comprehensibility' by two English language teachers on a scale of 0-5 based using a rating rubric prescribed by NAEP~\cite{naep} that spans the range from word-by-word reading to fluent reading with expression. With an inter-rater agreement of 0.76, we combine Z-score normalised scores of the two raters to derive the final target ratings. We provide this rich set of continuous normalised ratings as ground truth for training our systems. 

We evaluate our system using two measures that are motivated for fluency prediction on a continuous scale. The Pearson correlation measures the strength of the linear correlation between the predicted and target scores and is designed to be scale-free and bias-free. However given that our rating levels are associated with specific stages of fluency acquisition, inter-rater agreement must also be considered.  The Concordance Correlation Coefficient (CCC)~\cite{ccc_original} is such a measure that balances correlation with agreement. 

\section{SYSTEM ARCHITECTURES}

We present the wav2vec model based systems which are experimentally evaluated on our task. We first consider an architecture where simple pooling across the utterance of wav2vec frame-level features serves as the representation for the utterance comprehensibility prediction. The second architecture is motivated by the importance of the word as a unit in prosody representations. We evaluate different publicly available pre-trained wav2vec models and also investigate the importance of the different transformer layers to our task via layer weighting.

\subsection{Wav2vec2.0 Based Architectures}
Wav2vec2.0~\cite{wav2vec} is a self-supervised model pre-trained on a large speech corpus. It consists of a convolutional neural network (CNN) and a transformer. 
The entire model (CNN and transformer) is trained end-to-end by masking some of the latent representations and asking the transformer to predict their quantised versions from the neighbouring contexts. 
Since this pre-training procedure is task-agnostic (not tuned for any particular downstream task), the model learns a robust representation of speech at the frame-level given the vast unlabeled speech data that is typically used. The model's embeddings for any new speech input are used as features for given downstream tasks such as ASR, emotion recognition and speaker verification, possibly after a stage of fine-tuning on the task itself.  The original Wav2vec2.0~\cite{wav2vec} paper reported competitive performance for ASR on LibriSpeech by using just 10 minutes of labelled data. A simple linear projection layer is added on top of the wav2vec embeddings to predict phones and the model is fine-tuned using CTC loss. The simplicity of the projection layer hints towards the information-rich nature of the wav2vec representations. Given that fluency perception is influenced by both the segmental and suprasegmental attributes of speech, it is of interest to study the model for reading fluency prediction via the proposed architectures presented here.


\subsubsection{W2Vanilla}

\begin{figure}[tb]
    \centering
    \includegraphics[scale = 0.45]{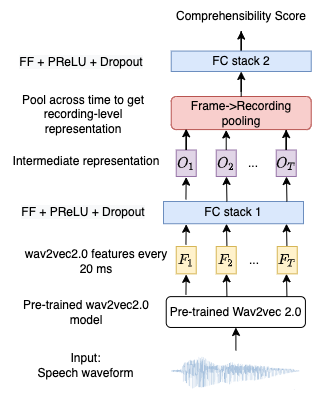}
    \caption{W2Vanilla Architecture}
    \label{fig:wav2}
\end{figure}

Our first wav2vec-based architecture (called Vanilla Wav2vec2.0 or W2Vanilla) uses the pre-trained wav2vec frame-level embeddings only as shown in Fig. \ref{fig:wav2}. We pass them through a stack of fully-connected (FC) layers and then mean-pool across the utterance to get a single utterance-level embedding. Each layer in the stack of fully connected layers consists of a fully-connected layer, an activation function (PReLU in our case) and dropout. The pooled embedding is passed through another stack of FC layers with one output neuron in the final layer, which on Sigmoid activation, can be interpreted as the comprehensibility score. Key hyperparameters of this architecture are the two FC stacks, whose depth and number of hidden units in each hidden layer.
\subsubsection{W2VAligned} \label{sssec:w2aligned}
\begin{figure}[tb]
    \centering
    \includegraphics[scale = 0.5]{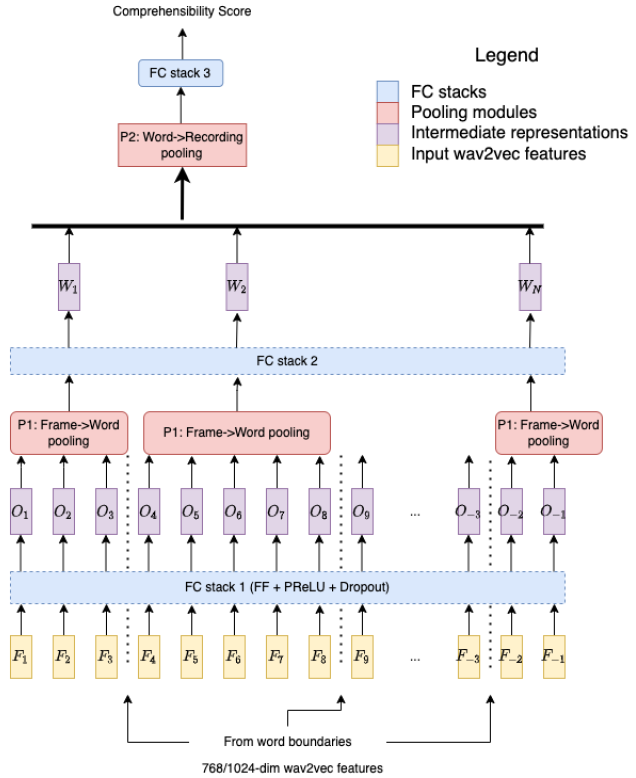}
    \caption{W2VAligned Architecture}
    \vspace{-7mm}
    \label{fig:wav3}
\end{figure}

The utterance level pooling of features potentially suppresses the across-word acoustic-prosodic variations considered important to fluency perception. We therefore introduce a new intermediate hierarchical pooling stage at the word level. Word boundaries are obtained by force-aligning the text hypothesis obtained from either an ASR decoder or the manually transcript. The wav2vec embeddings are then pooled for each word separately to get a word-level representation. For inter-word pauses, we can include either the pause before or after the word in it’s pooled representation. We choose to include the pause after the word since it can assist the model in detecting phrase breaks, an important prosodic event~\cite{ped_vaidya}. To reduce the number of trainable parameters, we use non-parametric techniques such as mean pooling across the word. The word-level representations are passed through a stack of FC layers that can extract contextual information across words. Post this, a second stage of pooling is applied to obtain a recording-level representation as depicted in Fig. \ref{fig:wav3}.

\subsection{Pre-trained Wav2vec2.0 Models}
Since performance of deep learning architectures scales with the amount of data, models pre-trained on large corpora have proven to be very effective, especially in low-resource scenarios. Wav2vec2.0~\cite{wav2vec} architecture was developed to train representations independent of target task specifications by asking the model to predict some of the masked latent representations from the unlabelled audio data itself. Our dataset consists of a total of 10 hours duration of audio data which is seemingly very less for pre-training a wav2vec model. To overcome this, we investigate the different pre-trained models available on Huggingface \cite{hf}, each model differing across the dataset it was pre-trained on and the training methodology.
\begin{itemize}
\item \textit{wav2vec2-base-960h} and \textit{wav2vec-large-960h} are pre-trained on LibriSpeech\cite{7178964} dataset which consists of 960 hours of total audio duration, sampled at 16kHz. The LibriSpeech data corpus contains audiobooks that are part of LibriVox project~\cite{kearns2014librivox} which is publicly available making it suitable for gathering unlabelled audio for the training of different tasks like speech recognition, emotion recognition, etc. The base/large model outputs representations of dimension 768/1024 respectively.
\item \textit{wav2vec2-large-xlsr-53} is trained using the same masked latent representation technique but for audio data comprising many different languages. The XLSR~\cite{conneau2020unsupervised} model is a wav2vec2.0 large model trained on Multilingual LibriSpeech\cite{pratap2020mls}, BABEL\cite{gales2014speech} and CommmonVoice\cite{ardila2019common} datasets which consists of 56k of hours of audio data and 53 different languages. 
\item \textit{wav2vec2-large-960h-lv60} is pre-trained using 960 hours of LibriSpeech\cite{7178964} and 60k of hours of Libri-Light\cite{9052942}.
\item \textit{{wav2vec2-large-960h-lv60-self}} uses the same dataset as the above model but the two differ in their training methodology. Instead of using the latent masked representation techniques, the model is trained with a self-training approach which uses pseudo-labels on the target task of automated speech recognition\cite{kahn2020self}. 
\end{itemize}

\subsection{Layer weighting}
The feature representations of pre-trained wav2vec2.0 models described above differ at two main aspects that affect our model architecture: Firstly the size of the transformer layers: base has a stack of 12 transformer layers while large has 24 layers stacked on top of each other. Secondly, base generates embeddings of dimension 768 while large generates 1024 dimensional embeddings.

The output of each layer of the transformer encoder is fed as input to the next layer. We can extract frame-level representations from either the last layer (output) or an intermediate layer. Previous works have demonstrated the effectiveness of using intermediate layer representations. For emotion recognition, experiments with different weighted combinations of the transformer layer outputs revealed the greater significance of intermediate layers rather than the output layer \cite{pepino2021emotion}. A work known as Audio ALBERT evaluated speaker verification and phoneme classification as a function of the layer and reported different layers as being optimal for different tasks\cite{aalbert}.

We would like to explore which layers are most informative for the fluency prediction task. However, optimising the layer weights jointly is computationally very intensive.  
Instead, we evaluate each layer individually and then further explore two weighting schemes as follows:
\begin{itemize}
    \item Mean pooling: at each time-step, take the average of the outputs of each layer of the transformer encoder to obtain the final frame-level embedding.
    \item Gaussian pooling: inspired by the weights learnt in \cite{pepino2021emotion}, we use a Gaussian distribution centred at the middle (e.g. between layer 12 and layer 13 for wav2vec2-large). The weights are normalised so that they sum to 1. The variance controls how rapidly the weights decay as we move away from the middle layer.
\end{itemize}


\begin{figure}[tb]
    \centering
    \includegraphics[scale = 0.5]{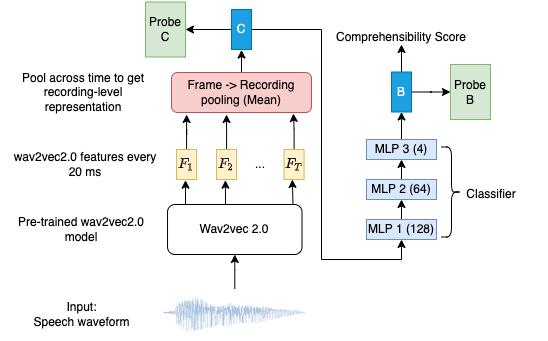}
    \caption{Location of probes in the Vanilla architecture. C is obtained on mean pooling the frame-level representations extracted from a pre-trained (frozen) wav2vec model. On passing it through 3 hidden layers with [128, 64, 4] hidden units, we get a compressed representation B $\in \mathbb{R}^4$. We regress the final score from this representation.}
    \label{fig:probe}
\end{figure}

\section{EXPERIMENTS and RESULTS} \label{sec:experiments}



The dataset of 1447 oral reading recordings is split into 6 speaker-independent folds with similar distribution of fluency scores across the folds. This is identical to the train-validate-test folds used in the previous work ~\cite{sabu2024predicting, kamini_thesis} allowing us to directly compare the performance obtained by the deep learning systems of this work with the previously reported performance from classification using knowledge-based lexical and acoustic-prosodic features. Further, the training loss function is chosen to be the Concordance Correlation Loss rather than the usual Mean Square Error (MSE) ~\cite{pandit2019many} ~\cite{atmaja2021evaluation}. 
The Concordance Correlation Loss (CCL = 1 - CCC) optimises the CCC which, as mentioned in \ref{sec:dataset}, is a more relevant evaluation metric for the fluency prediction task. 

We report the fluency prediction performance across our systems with the different system architectures and wav2vec models.  We also present an experiment probing the embeddings in one of the systems for selected knowledge-based features that have been shown to contribute to perceived fluency in the previous work~\cite{sabu2024predicting, kamini_thesis}. 

\subsection{W2Vanilla and W2VAligned}

Table \ref{tab:vanilla} reports the performance of W2Vanilla architecture with \textit{wav2vec2-large-960h} pre-trained model on the task of comprehensibility rating. This table also shows the performance of the architecture on selected individual intermediate layer embeddings as well as different layer combinations. As can be seen in table \ref{tab:vanilla}, individually, the layer 17 embedding is most predictive overall of utterance fluency, more so than the final output layer of Wav2vec2.0 that has traditionally worked well in ASR tasks~\cite{wav2vec}. From our experiments, we also note that mean and Gaussian combinations of intermediate layer embeddings outperform individual layer embeddings. Finally, and most importantly, the W2Vanilla model's performance(0.827 in table \ref{tab:vanilla}) already surpasses the knowledge based features system (0.794 in table \ref{tab:vanilla}) which uses classical Random Forest Regression for the fluency score prediction \cite{kamini_thesis,sabu2024predicting}. This is particularly surprising in view of the fact that the classical system assumes a knowledge of the canonical text while computing features that summarise the speaker's word-decoding errors and prosodic miscues, while this information is not explicitly available in the deep learning systems.

\begin{table}[tb]
    \begin{center}
    \caption{W2Vanilla architecture with \textit{wav2vec2-large-960h} pretrained model and different layer weighting and its comparison with W2VAligned and the hand-crafted features RFC model\cite{kamini_thesis,sabu2024predicting}}
    \begin{tabular}{|c|c|c|c|}
    \hline
    Which Layer & Val CCC & Test CCC & Test Pearson\\
    \hline
    1&0.723&0.696&0.721\\
    5&0.732&0.558&0.644\\
    10&0.803&0.803&0.813\\
    15&0.813&0.803&0.816\\
    16&0.814&0.809&0.817\\
    \textbf{17}&\textbf{0.816}&\textbf{0.808}&\textbf{0.821}\\
    18&0.809&0.808&0.817\\
    21&0.787&0.787&0.796\\
    \hline
    Output(24)&0.770&0.772&0.778\\
    Mean&0.804&0.808&0.825\\
    \textbf{Gaussian}&\textbf{0.805}&\textbf{0.808}&\textbf{0.827}\\
    \hline
    W2VAligned FA&0.798&0.802&0.821\\
    \hline
    \textbf{RFC}\cite{kamini_thesis,sabu2024predicting} & - & - & \textbf{0.794}\\
    \hline
    \end{tabular}
    \label{tab:vanilla}
    \end{center}
\end{table}

In table \ref{tab:vanilla}, the performance of W2VAligned architecture is also reported which uses manually transcribed text(FA) for creating alignments as explained in section \ref{sssec:w2aligned}. Contrary to our expectations, performance slightly reduces on introducing this intermediate pooling stage. Explanations for this could be the more complex architecture, a rigid pooling mechanism, coupled with possibly erroneous word boundary alignments. 


\subsection{Pre-trained Model Embeddings}

Table \ref{tab:pre} reports the performance comparison of different wav2vec pre-trained models which are used for extraction of audio embeddings that are later used for comprehensibility prediction task. From the table, we can infer \textit{xlsr} model performs worse as multi-lingual pre-training is leading to loss of useful features for English language. \textit{wav2vec2-large-lv60-self}, which uses self-training approach with ASR as target task for pre-training performs the best among all pre-trained models. The better performance of pre-trained model mentioned here may be linked to ASR task-oriented pre-training of the model with self-training approach. Since the hyperparameters for W2Vanilla architecture for the above experiment were tuned on \textit{wav2vec2-large-960h} model, further improvement in performance is expected by separately tuning the hyperparameters for each of the pre-trained models.

\begin{table}[tb]
    \begin{center}
    \caption{Different pre-trained models on W2Vanilla architecture}
    \begin{tabular}{|c|c|c|c|c|}
    \hline
      \shortstack{Pre-trained\\Model}&\shortstack{Layer\\Weighting}&\shortstack{Val\\CCC}&\shortstack{Test\\CCC}&\shortstack{Test\\Pearson}\\
    \hline
    wav2vec2-base-&Mean&0.776&0.787&0.808\\
    960h&Gaussian&0.796&0.797&0.816\\
    \hline
    wav2vec2-large-&Mean&0.804&0.808&0.825\\
    960h&Gaussian&0.805&0.808&0.827\\
    \hline
    wav2vec2-large-&Mean&0.783&0.777&0.805\\
    960h-lv60&Gaussian&0.790&0.790&0.813\\
    \hline
    wav2vec2-large-&Mean&0.759&0.762&0.797\\
    xlsr-53&Gaussian&0.799&0.775&0.797\\
    \hline
    wav2vec2-large-&Mean&0.807&0.810&0.827\\
    960h-lv60-self&Gaussian&\textbf{0.811}&\textbf{0.814}&\textbf{0.831}\\
    \hline
    \end{tabular}
    \label{tab:pre}
    \end{center}
\end{table}

\subsection{Probing the representations}

\begin{figure}
    \centering
    \includegraphics[width=\columnwidth]{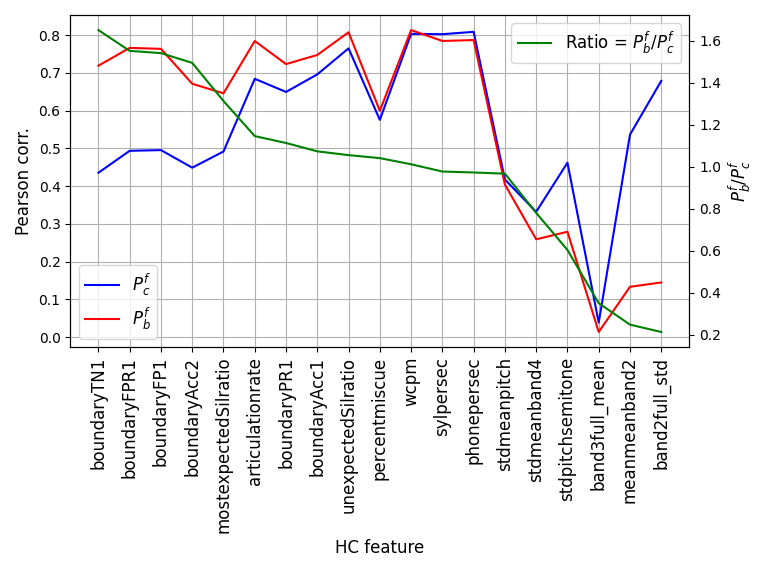}
    \caption{$P^f_c$ (Performance on wav2vec embedding), $P^f_b$ (Performance on bottleneck embedding) and the ratio 
    sorted in descending order according to the ratio. The HC (hand-crafted) features are taken from ~\cite{kamini_thesis}. }
    \label{fig:probe_res}
\end{figure}

Given the competitive prediction performances of the deep learning architectures, we probe the pooled wav2vec representation (C in Fig.~\ref{fig:probe}) to understand which knowledge-based features might be captured by the pre-trained wav2vec model. We also train the classifier layers using the comprehensibility scores and probe the final hidden layer embedding (B in Fig. ~\ref{fig:probe}). Since B is a compressed representation, only those features which are crucial for fluency will be preserved through the bottleneck classifier layers. We use the following methodology for probing:

Let X be be the input embedding matrix (of dimensions d x n) where d is the dimension of the embedding (1024 for C and 4 for B) and n is the number of recordings in the dataset. $Y^{f} \in \mathbb{R}^{n}$ is the concatenation of feature f across the dataset. We split the dataset into speaker non-overlapping train and test subsets (in the ratio 70-30). Using the train subset, we correlate each dimension of our embedding with feature f and the dimension with the highest correlation ($k^f$) is stored:
\[k^{f} = \underset{i=1,2,...,d}{\mathrm{argmax}}\, \, \, \, Corr(X_{train}[i, :], Y^{f}_{train})\]
We then correlate the same dimension k with the HC features for the test subset and report performance $P^f$:
\[P^{f} = Corr(X_{test}[k^{f}, :], Y^{f}_{test})\]
In both cases, $Corr(.\,, .)$ refers to the Pearson correlation between the two sequences. We compute $P^{f}$ for 19 hand-crafted features which were shown to be most helpful for comprehensibility prediction in~\cite{kamini_thesis, sabu2024predicting}.
We carry out the above procedure for both the wav2vec-pre trained embedding (C) and the bottleneck output (B). Let $P^{f}_c$ and $P^{f}_b$ be the respective Pearson correlations on the test subset. We also investigate the ratio $\frac{P^{f}_b}{P^{f}_b}$; higher values indicate the feature is important since it is preserved through the bottleneck. We chose this simple correlation-based technique over training a linear/MLP probe due to two reasons. Firstly, the probe requires extensive tuning in terms of complexity and hyperparameters~\cite{hewitt2019designing}. Secondly, for the wav2vec embedding with d = 1024, training a simple linear regression model leads to overfitting since number of samples in the train set n $<$ d.

Results on probing mean-pooled \textit{wav2vec2-large-960h} are presented in Fig.~\ref{fig:probe_res}. We observe that phrase boundary aggregates (\textit{boundaryAcc, boundaryFPR, boundaryTN}) and speech rate features (\textit{wcpm}, phones per second (\textit{phonepersec}) and syllables per second (\textit{sylpersec})) can be extracted from both B and C. 
In fact, we see higher performance for B, which indicates that the classifier layers perform additional feature learning to compute the phrase boundary aggregates. On the other hand, most AP contour aggregates such as mean/standard deviation of various bands (\textit{meanmeanband2, band2full\_std}) and pitch aggregates (\textit{stdpitchsemitone}) are captured by pre-trained wav2vec but they are not preserved through the bottleneck. 

\section{CONCLUSIONS} \label{sec:conclusions}
This work proposed the first end-to-end trained solution for oral reading fluency prediction. Features extracted from the waveform by a pre-trained wav2vec model turned out to be more informative on the same task and dataset than a previously published system using carefully hand-crafted features and a knowledge of the intended text~\cite{kamini_thesis,sabu2024predicting}. A study probing the learned embeddings revealed significant correlations with the high-level attributes of perceived fluency such as speech rate and prosodic miscues. While speech rate can understandably be represented by waveform acoustics, it is possible that this speaking behaviour is strongly correlated with the prosody quality in our data, leading to the latter being highly predictable from waveform embeddings. 
Future work will investigate the possibly complementary information of features poorly correlated with the embeddings via feature concatenation. Finally, the use of wav2vec models opens up the valuable prospect of benefitting from large amounts of unlabeled data of children's oral reading recordings by the self-supervised training.  


\bibliographystyle{IEEEtran}

\bibliography{conference_101719.bib}

\begin{thebibliography}{10}
\providecommand{\url}[1]{#1}
\csname url@samestyle\endcsname
\providecommand{\newblock}{\relax}
\providecommand{\bibinfo}[2]{#2}
\providecommand{\BIBentrySTDinterwordspacing}{\spaceskip=0pt\relax}
\providecommand{\BIBentryALTinterwordstretchfactor}{4}
\providecommand{\BIBentryALTinterwordspacing}{\spaceskip=\fontdimen2\font plus
\BIBentryALTinterwordstretchfactor\fontdimen3\font minus \fontdimen4\font\relax}
\providecommand{\BIBforeignlanguage}[2]{{%
\expandafter\ifx\csname l@#1\endcsname\relax
\typeout{** WARNING: IEEEtran.bst: No hyphenation pattern has been}%
\typeout{** loaded for the language `#1'. Using the pattern for}%
\typeout{** the default language instead.}%
\else
\language=\csname l@#1\endcsname
\fi
#2}}
\providecommand{\BIBdecl}{\relax}
\BIBdecl

\bibitem{aser}
A.~Centre, ``Aser: The annual status of education report (rural) 2016,'' Tech. Rep., 2017.

\bibitem{torgesen1998catch}
J.~Torgesen, ``Catch them before they fall: Identification and assessment to prevent reading failure in young children (on-line),'' \emph{National Institute of Child Health and Human Development. Available: ldonline. org. ld\_indepth/reading/torgesen\_catchthem. html}, pp. 1--15, 1998.

\bibitem{kuhn2010aligning}
M.~R. Kuhn, P.~J. Schwanenflugel, and E.~B. Meisinger, ``Aligning theory and assessment of reading fluency: Automaticity, prosody, and definitions of fluency,'' \emph{Reading research quarterly}, vol.~45, no.~2, pp. 230--251, 2010.

\bibitem{hasbrouck1992curriculum}
J.~E. Hasbrouck and G.~Tindal, ``Curriculum-based oral reading fluency norms for students in grades 2 through 5,'' \emph{Teaching Exceptional Children}, vol.~24, no.~3, pp. 41--44, 1992.

\bibitem{valencia2010oral}
S.~W. Valencia, A.~T. Smith, A.~M. Reece, M.~Li, K.~K. Wixson, and H.~Newman, ``Oral reading fluency assessment: Issues of construct, criterion, and consequential validity,'' \emph{Reading Research Quarterly}, vol.~45, no.~3, pp. 270--291, 2010.

\bibitem{kamini_thesis}
K.~Sabu, ``Automatic assessment of fluency in children’s oral reading using prosody modeling,'' Ph.D. dissertation, Indian Institute of Technology Bombay, July 2022.

\bibitem{sabu2024predicting}
K.~Sabu and P.~Rao, ``Predicting children’s perceived reading proficiency with prosody modeling,'' \emph{Computer Speech \& Language}, vol.~84, p. 101557, 2024.

\bibitem{yu2015using}
Z.~Yu, V.~Ramanarayanan, D.~Suendermann-Oeft, X.~Wang, K.~Zechner, L.~Chen, J.~Tao, A.~Ivanou, and Y.~Qian, ``Using bidirectional lstm recurrent neural networks to learn high-level abstractions of sequential features for automated scoring of non-native spontaneous speech,'' in \emph{2015 IEEE Workshop on Automatic Speech Recognition and Understanding (ASRU)}.\hskip 1em plus 0.5em minus 0.4em\relax IEEE, 2015, pp. 338--345.

\bibitem{chen2018end}
L.~Chen, J.~Tao, S.~Ghaffarzadegan, and Y.~Qian, ``End-to-end neural network based automated speech scoring,'' in \emph{2018 IEEE International Conference on Acoustics, Speech and Signal Processing (ICASSP)}.\hskip 1em plus 0.5em minus 0.4em\relax IEEE, 2018, pp. 6234--6238.

\bibitem{grover}
\BIBentryALTinterwordspacing
M.~S. Grover, Y.~Kumar, S.~Sarin, P.~Vafaee, M.~Hama, and R.~R. Shah, ``Multi-modal automated speech scoring using attention fusion,'' 2020. [Online]. Available: \url{https://arxiv.org/abs/2005.08182}
\BIBentrySTDinterwordspacing

\bibitem{qian_scoring}
Y.~Qian, P.~Lange, K.~Evanini, R.~Pugh, R.~Ubale, M.~Mulholland, and X.~Wang, ``Neural approaches to automated speech scoring of monologue and dialogue responses,'' in \emph{ICASSP 2019 - 2019 IEEE International Conference on Acoustics, Speech and Signal Processing (ICASSP)}, 2019, pp. 8112--8116.

\bibitem{zechner2009automatic}
K.~Zechner, D.~Higgins, X.~Xi, and D.~M. Williamson, ``Automatic scoring of non-native spontaneous speech in tests of spoken english,'' \emph{Speech Communication}, vol.~51, no.~10, pp. 883--895, 2009.

\bibitem{wav2vec}
A.~Baevski, Y.~Zhou, A.~Mohamed, and M.~Auli, ``wav2vec 2.0: A framework for self-supervised learning of speech representations,'' \emph{Advances in Neural Information Processing Systems}, vol.~33, pp. 12\,449--12\,460, 2020.

\bibitem{pasad2021layer}
A.~Pasad, J.-C. Chou, and K.~Livescu, ``Layer-wise analysis of a self-supervised speech representation model,'' in \emph{2021 IEEE Automatic Speech Recognition and Understanding Workshop (ASRU)}.\hskip 1em plus 0.5em minus 0.4em\relax IEEE, 2021, pp. 914--921.

\bibitem{shah2021all}
J.~Shah, Y.~K. Singla, C.~Chen, and R.~R. Shah, ``What all do audio transformer models hear? probing acoustic representations for language delivery and its structure,'' \emph{arXiv preprint arXiv:2101.00387}, 2021.

\bibitem{naep}
S.~White, J.~Sabatini, B.~J. Park, J.~Chen, J.~Bernstein, and M.~Li, ``The 2018 naep oral reading fluency study. nces 2021-025.'' \emph{National Center for Education Statistics}, 2021.

\bibitem{ccc_original}
I.~Lawrence and K.~Lin, ``A concordance correlation coefficient to evaluate reproducibility,'' \emph{Biometrics}, pp. 255--268, 1989.

\bibitem{ped_vaidya}
M.~Vaidya, K.~Sabu, and P.~Rao, ``Deep learning for prominence detection in children’s read speech,'' in \emph{ICASSP 2022 - 2022 IEEE International Conference on Acoustics, Speech and Signal Processing (ICASSP)}, 2022, pp. 8157--8161.

\bibitem{hf}
``Hugging face,'' \url{https://huggingface.co/facebook}, accessed: August 2022.

\bibitem{7178964}
V.~Panayotov, G.~Chen, D.~Povey, and S.~Khudanpur, ``Librispeech: An asr corpus based on public domain audio books,'' in \emph{2015 IEEE International Conference on Acoustics, Speech and Signal Processing (ICASSP)}, April 2015, pp. 5206--5210.

\bibitem{kearns2014librivox}
J.~Kearns, ``Librivox: Free public domain audiobooks,'' \emph{Reference Reviews}, 2014.

\bibitem{conneau2020unsupervised}
A.~Conneau, A.~Baevski, R.~Collobert, A.~Mohamed, and M.~Auli, ``Unsupervised cross-lingual representation learning for speech recognition,'' \emph{arXiv preprint arXiv:2006.13979}, 2020.

\bibitem{pratap2020mls}
V.~Pratap, Q.~Xu, A.~Sriram, G.~Synnaeve, and R.~Collobert, ``Mls: A large-scale multilingual dataset for speech research,'' \emph{arXiv preprint arXiv:2012.03411}, 2020.

\bibitem{gales2014speech}
M.~J. Gales, K.~M. Knill, A.~Ragni, and S.~P. Rath, ``Speech recognition and keyword spotting for low-resource languages: Babel project research at cued,'' in \emph{Fourth International workshop on spoken language technologies for under-resourced languages (SLTU-2014)}.\hskip 1em plus 0.5em minus 0.4em\relax International Speech Communication Association (ISCA), 2014, pp. 16--23.

\bibitem{ardila2019common}
R.~Ardila, M.~Branson, K.~Davis, M.~Henretty, M.~Kohler, J.~Meyer, R.~Morais, L.~Saunders, F.~M. Tyers, and G.~Weber, ``Common voice: A massively-multilingual speech corpus,'' \emph{arXiv preprint arXiv:1912.06670}, 2019.

\bibitem{9052942}
J.~Kahn, M.~Rivière, W.~Zheng, E.~Kharitonov, Q.~Xu, P.~Mazaré, J.~Karadayi, V.~Liptchinsky, R.~Collobert, C.~Fuegen, T.~Likhomanenko, G.~Synnaeve, A.~Joulin, A.~Mohamed, and E.~Dupoux, ``Libri-light: A benchmark for asr with limited or no supervision,'' in \emph{ICASSP 2020 - 2020 IEEE International Conference on Acoustics, Speech and Signal Processing (ICASSP)}, 2020, pp. 7669--7673.

\bibitem{kahn2020self}
J.~Kahn, A.~Lee, and A.~Hannun, ``Self-training for end-to-end speech recognition,'' in \emph{ICASSP 2020-2020 IEEE International Conference on Acoustics, Speech and Signal Processing (ICASSP)}.\hskip 1em plus 0.5em minus 0.4em\relax IEEE, 2020, pp. 7084--7088.

\bibitem{pepino2021emotion}
L.~Pepino, P.~Riera, and L.~Ferrer, ``Emotion recognition from speech using wav2vec 2.0 embeddings,'' \emph{arXiv preprint arXiv:2104.03502}, 2021.

\bibitem{aalbert}
P.-H. Chi, P.-H. Chung, T.-H. Wu, C.-C. Hsieh, Y.-H. Chen, S.-W. Li, and H.-y. Lee, ``Audio albert: A lite bert for self-supervised learning of audio representation,'' in \emph{2021 IEEE Spoken Language Technology Workshop (SLT)}.\hskip 1em plus 0.5em minus 0.4em\relax IEEE, 2021, pp. 344--350.

\bibitem{pandit2019many}
V.~Pandit and B.~Schuller, ``The many-to-many mapping between the concordance correlation coefficient and the mean square error,'' \emph{arXiv preprint arXiv:1902.05180}, 2019.

\bibitem{atmaja2021evaluation}
B.~T. Atmaja and M.~Akagi, ``Evaluation of error-and correlation-based loss functions for multitask learning dimensional speech emotion recognition,'' in \emph{Journal of Physics: Conference Series}, vol. 1896, no.~1.\hskip 1em plus 0.5em minus 0.4em\relax IOP Publishing, 2021, p. 012004.

\bibitem{hewitt2019designing}
J.~Hewitt and P.~Liang, ``Designing and interpreting probes with control tasks,'' \emph{Proceedings of the 2019 Con}, 2019.

\end{thebibliography}


\end{document}